\def\year{2022}\relax
\documentclass[letterpaper]{article} % DO NOT CHANGE THIS
\usepackage{aaai22}  % DO NOT CHANGE THIS
\usepackage{times}  % DO NOT CHANGE THIS
\usepackage{helvet}  % DO NOT CHANGE THIS
\usepackage{courier}  % DO NOT CHANGE THIS
\usepackage[hyphens]{url}  % DO NOT CHANGE THIS
\usepackage{graphicx} % DO NOT CHANGE THIS
\usepackage{natbib}  % DO NOT CHANGE THIS AND DO NOT ADD ANY OPTIONS TO IT
\usepackage{caption} % DO NOT CHANGE THIS AND DO NOT ADD ANY OPTIONS TO IT
\DeclareCaptionStyle{ruled}{labelfont=normalfont,labelsep=colon,strut=off} % DO NOT CHANGE THIS
\usepackage{algorithm}
\usepackage{algorithmic}
\usepackage{array}
\usepackage{multirow}
\usepackage{subfigure}
\usepackage[table ]{ xcolor}
\usepackage{amsmath}
\usepackage{amssymb}

\newcommand{\PreserveBackslash}[1]{\let\temp=\\#1\let\\=\temp}

\newcolumntype{C}[1]{>{\PreserveBackslash\centering}p{#1}}

\def\eg{\emph{e.g}.} 
\def\ie{\emph{i.e}.} 
 
\def\etc{\emph{etc}.} 
 
\def\etal{\emph{et al}.}

\definecolor{mygray}{RGB}{230,230,230}
\definecolor{mygray2}{RGB}{130,130,130}
\newcommand{\red}[1]{{\color{red}{#1}}}
\newcommand{\blue}[1]{{\color{blue}{#1}}}
\newcommand{\green}[1]{{\color{green}{#1}}}
\newcommand{\yellow}[1]{{\color{yellow}{#1}}}
\usepackage{newfloat}
\usepackage{listings}
\floatstyle{ruled}
\newfloat{listing}{tb}{lst}{}
\floatname{listing}{Listing}
\title{ACGNet: Action Complement Graph Network\\for Weakly-supervised Temporal Action Localization}
\author{
    Zichen Yang\textsuperscript{\rm 1},
    Jie Qin\textsuperscript{\rm 2},
	Di Huang\textsuperscript{\rm 1}\thanks{indicates the corresponding author.}
}

\affiliations {
    \textsuperscript{\rm 1} State Key Laboratory of Software Development Environment, Beihang University, Beijing 100191, China\\
    \textsuperscript{\rm 2} College of Computer Science and Technology, Nanjing University of Aeronautics and Astronautics, Nanjing 211106, China\\
    \{yangzichen, dhuang\}@buaa.edu.cn, %dhuang@buaa.edu.cn,
    qinjiebuaa@gmail.com
}

\usepackage{bibentry}
\begin{document}

\maketitle

\begin{abstract}
Weakly-supervised temporal action localization (WTAL) in untrimmed videos has emerged as a practical but challenging task since only video-level labels are available. Existing approaches typically leverage off-the-shelf segment-level features, which suffer from spatial incompleteness and temporal incoherence, thus limiting their performance. In this paper, we tackle this problem from a new perspective by enhancing segment-level representations with a simple yet effective graph convolutional network, namely action complement graph network (ACGNet). It facilitates the current video segment to perceive spatial-temporal dependencies from others that potentially convey complementary clues, implicitly mitigating the negative effects caused by the two issues above. By this means, the segment-level features are more discriminative and robust to spatial-temporal variations, contributing to higher localization accuracies. More importantly, the proposed ACGNet works as a universal module that can be flexibly plugged into different WTAL frameworks, while maintaining the end-to-end training fashion. Extensive experiments are conducted on the THUMOS'14 and ActivityNet1.2 benchmarks, where the state-of-the-art results clearly demonstrate the superiority of the proposed approach.
\end{abstract}

%%%%%%%%% BODY TEXT
\section{Introduction}

Understanding human actions in videos is an important research direction and has been actively studied in the computer vision community \cite{LFB, BCN, ECO, zsecoc, TEA, stagnet, TEINet, SlowFast, sports, icassp, igmn}. The fundamental step is to build meaningful spatial-temporal representations, which involve not only static features from each frame, but also dynamic dependencies across consecutive frames. Among the main tasks in action understanding, temporal action localization \cite{CRCNN,BSN,BMN} has received tremendous efforts in the past several years, with a wide range of applications (\eg, intelligent surveillance, video retrieval, and human-computer interaction).

\begin{figure}[!t]
\begin{center}
\includegraphics[width=0.99\linewidth]{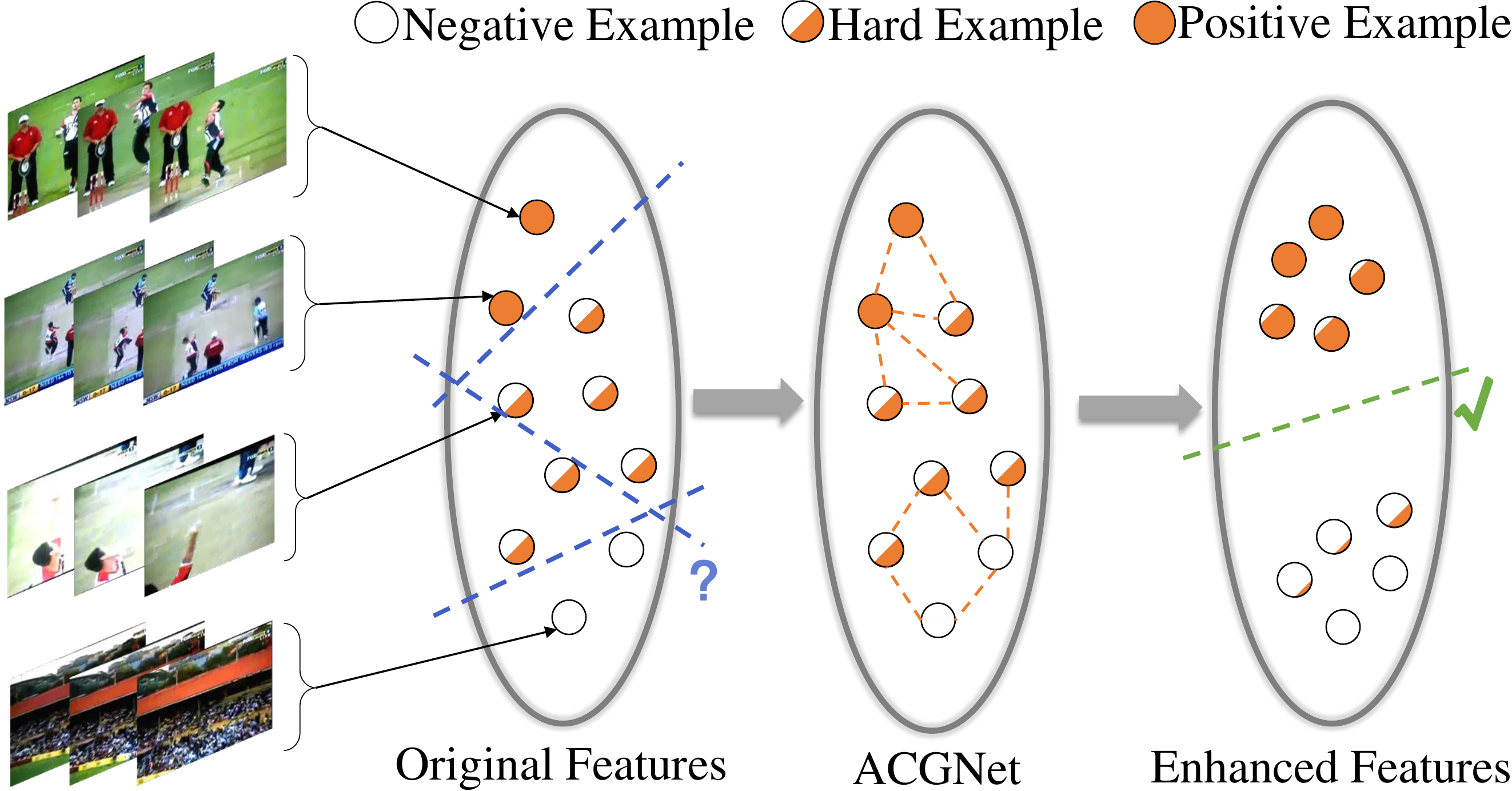}
\end{center}
   \caption{Intuition behind the proposed action complement graph network (ACGNet). By exploiting complementary information across different segments, more discriminative segment-level action representations are learned, leading to more accurate localization results. The blue/green dashed lines indicate the classification hyperplanes.}
\label{learning}
\end{figure}

To achieve accurate localization results, conventional (fully-supervised) temporal action localization (FTAL) methods \cite{CDC,BSN,BMN,SSN,STEP} often make use of deep convolutional neural networks (CNNs) trained on video datasets with frame-level annotations. Unfortunately, as the dataset size grows rapidly and the total video length even reaches several decades \cite{youtube}, it is obviously unrealistic to acquire such fine-grained annotations. To this end, weakly-supervised temporal action localization (WTAL) \cite{UntrimmedNets}, where only video-level action categories are annotated, has recently emerged as a more practical task. To tackle WTAL, a common practice is to uniformly sample short segments of equal length, for which classifiers are trained (usually through multiple instance learning \cite{WTALC}) with video-level labels, and localization results are generated based on the classification/activation scores of each segment with regard to action categories.

However, in this paradigm, the evenly sampling strategy incurs two critical issues that greatly limit localization performance.
%As shown in Figure~\ref{overview}, 
On the one hand, the action segments often suffer from occlusion, blurring, out of field, \emph{etc}., thus lack of certain spatial details. On the other hand, a complete action usually spans a long temporal window and a short action segment is insufficient to observe the full dynamics of that action. We respectively identify the two issues as `spatial incompleteness' and `temporal incoherence' of an action segment, both of which make predictions in WTAL unreliable.

In this work, we implicitly address the two issues by a simple yet effective graph convolutional network. The proposed action complement graph network (ACGNet) facilitates an action segment to exploit complementary clues from other segments across the entire untrimmed long video. As shown in Figure~\ref{learning}, after applying our ACGNet, those hard examples can be more easily classified based on the enhanced features. Specifically, we not only consider segment-level similarities but also mitigate negative influences of temporally close segments when constructing the initial action complement graph (ACG). Besides, we make this graph sparse enough to preserve the most informative connections. Through graph convolutions, the complementary information from high-quality segments is propagated to low-quality ones, leading to the enhanced action representation for each segment. In other words, the complementary information provided by other segments is regarded as supervision to learn more discriminative features in the WTAL scenario. Most importantly, owing to the delicately-designed loss function, our ACGNet works as a generic plug-in module and can be flexibly embedded into different WTAL frameworks, further remarkably strengthening the state-of-the-art performance. 

In summary, our main contributions are three-fold:
\begin{itemize}
    \item We propose a novel graph convolutional network for WTAL, namely ACGNet, which greatly enhances the discriminability of segment-level action representations by implicitly exploiting the complementary information and jointly addressing the issues of spatial incompleteness and temporal incoherence.
    \item We consider multiple vital factors (\ie, segment similarity, temporal diffusion, and graph sparsity) to construct the initial ACG. Moreover, we make the training of our graph network feasible and practical by proposing a novel `easy positive mining' loss, endowing our ACGNet with the flexibility to be injected into existing frameworks without bells and whistles.
    \item We equip several recent WTAL methods with the proposed ACGNet. Extensive experiments on two challenging datasets demonstrate its capability to further push the state of the art in WTAL to a large extent.
\end{itemize}

\section{Related Work}
\noindent\textbf{Fully-supervised Temporal Action Localization}.
Action localization has recently attracted numerous research interests \cite{zhang2019adversarial,Victor2016DAPs,LinSingle,BSN,li2019long}. A typical pipeline is to first generate temporal action proposals and then classify pre-defined actions based on the proposals. For example, \cite{CDC} proposes a Convolutional-De-Convolutional filter through temporal upsampling and spatial downsampling to precisely detect segment boundaries. \cite{SSN} presents the Structured Segment Network to model the temporal structure of each action segment via a structured temporal pyramid. \cite{STEP} provides an end-to-end progressive optimization framework (STEP) for more effective spatial-temporal modeling.

\begin{figure*}[!t]
\begin{center}
\includegraphics[width=0.99\linewidth]{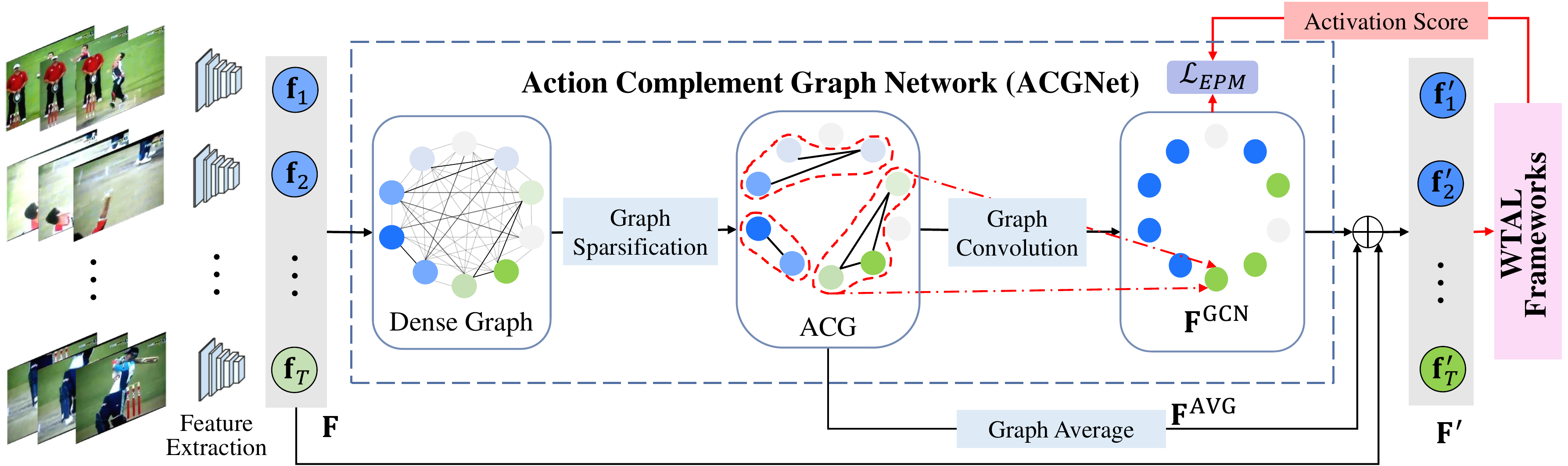}
\end{center}
   \caption{Overall framework of the proposed ACGNet, which takes the segment-level features as input and generates enhanced, more discriminative features by exploiting complementary clues between different segments. More importantly, our ACGNet can be flexibly plugged into various existing WTAL frameworks without bells and whistles.}
\label{overview}
\end{figure*}

\noindent\textbf{Weakly-supervised Temporal Action Localization}.
Regarding WTAL, only category labels for whole videos are available, without any fine-grained annotation for each action instance. To tackle this challenge, existing methods usually segment the video at equal temporal intervals, and then classify each segment by multiple instance learning. Specifically, the activation score of a segment to each category, \ie, class activation sequence (CAS), is calculated to classify the action segment. \cite{UntrimmedNets} formally proposes the tasks of `weakly supervised action recognition and temporal localization' and used attention weights to exclude the video clips that do not contain actions. \cite{BaSNet} presents BaS-Net by introducing a background class to assist training, inhibiting the activation of background frames to improve the positioning performance. \cite{DGAM} deliver a frame-level probability distribution model (\ie, DGAM) based on frame-level attention to distinguish action frames from background frames. BaM \cite{BaM} is an improved variant of BaS-Net, which employs multiple instance learning to estimate the uncertainty of video frame classification and model the background frames.

Currently, most of the existing studies focus more on developing various learning techniques to improve localization performance based on pre-extracted segment-level features. By contrast, this work emphasizes on the enhancement of the segment-level features by exploring and exploiting the spatial-temporal complementarity across segments, universally benefiting different WTAL frameworks.

\noindent\textbf{Graph-based Temporal Action Localization}.
Recently, some works investigate graph learning to fuse the information among related categories, multiple proposals or multiple sub-actions to infer the possible actions of a certain segment. For example, P-GCN \cite{PGCN} constructs a graph according to the distances and IoUs between proposals, aiming to adjust the category and boundary of each proposal by using context information. G-TAD \cite{G_TAD} attempts to make use of not only temporal context, but also semantic context captured through graph convolutional networks (GCN), and then temporal action detection is cast as a sub-graph localization problem. GTRM \cite{GTRM} employs GCN to integrate all the action segments within a certain period of time in the action segmentation task. All such efforts are made in the fully-supervised setting.

In WTAL, \cite{Action_Graph} establishes a similarity graph to understand how an action appears as well as the sub-actions that comprise the action's full extent. Notably, this is essentially different from our purpose to complement and enhance features by fully mining the complementary information across segments. Moreover, they design a fixed WTAL network, while our ACGNet works as a universal module to improve various WTAL frameworks. In addition, we propose different graph designs and a novel loss function that enables the joint training of ACGNet and WTAL frameworks.

\section{Action Complement Graph Network}

As mentioned above, an input video is uniformly divided into multiple temporal segments, based on which WTAL is performed. The localization accuracy highly depends on the discriminability of segment-level action representations, especially in our weakly-supervised setting. To this end, we aim to enhance segment-level representations, by exploiting the complementary information among different segments. Since our ACGNet is essentially designed for feature enhancement, it can be flexibly plugged into various existing WTAL frameworks, such as \cite{BaSNet,BaM,DGAM} used in our experiments. In the following, we first give a brief introduction of the entire proposed network. Subsequently, we elaborate how to construct the action complement graph (ACG) in a principled way, and how to enhance features based on graph convolution, respectively. Finally, a novel loss is presented to make the training of our graph network feasible. After embedding the ACGNet into existing WTAL frameworks, we follow the standard pipeline provided in \cite{BaSNet,BaM,DGAM} to generate the final localization results.

\subsection{Method Overview}
\label{overview1}

Figure \ref{overview} illustrates the overall framework of the proposed ACGNet. Given an input video $V$, we first evenly divide it into a fixed number of $T$ short temporal segments $\{S_t\}_{t=1}^T$, to handle large variations in video lengths. Then, we extract the features of these segments by using a widely-adopted video feature extraction network, \eg, the I3D network~\cite{I3D}. The extracted segment-level features are denoted by the $D$-dimensional feature vectors $\mathbf{f}_t \in \mathbb{R}^D$, which can be concatenated to form the video-level representation $\mathbf{F}=[\mathbf{f}_1,~\mathbf{f}_2,~\cdots,~\mathbf{f}_T] \in \mathbb{R}^{T \times D}$. 

The proposed ACGNet receives the original features $\mathbf{F}$ as input and generates the enhanced features $\mathbf{F}'$ based on a graph convolutional network. The action complement graph (ACG) is constructed for each video in a principled way to exchange complementary information between its nodes (\ie, segments). After constructing the ACG, node-level features are propagated and fused by using graph convolution operations. The output graph features can be regarded an enhanced and complementary counterpart of the original features. Finally, the original and the enhanced features are combined as the ultimate discriminative features $\mathbf{F}'$, which can be used as the input to any WTAL methods to improve their localization performance to a large extent. In addition, a novel loss is proposed to facilitate the joint training of our ACGNet and existing WTAL frameworks.

\subsection{Action Complement Graph}
\label{acg}
Due to the lack of frame-level annotations, it is difficult to classify individual short segments. However, multiple segments (among which there usually exist easy-to-classify action instances) in a video can complement each other. Thereby, the ACG is to capture the complementary relationships and enhance the representation for each segment.

Formally, the ACG is defined as $\mathcal{G}=(\mathcal{V},\mathcal{E})$. $\mathcal{V}$ denotes a set of nodes $\{v_t\}_{t=1}^{T}$, corresponding to $T$ segment-level features $\{\mathbf{f}_t\}_{t=1}^{T}$, while $\mathcal{E}$ refers to the edge set where $e_{ij}=(v_i, v_j)$ is the edge between the nodes $v_i$ and $v_j$. In addition, we define $\mathbf{A}\in \mathbb{R}^{T\times T}$ as the adjacency matrix associated with the graph $\mathcal{G}$. The weight of an edge, \ie $A_{ij}$, represents the strength of the relationship between two connected nodes, and a larger weight indicates that two segments are more associated to each other.

In the subsequent, we introduce how to construct ACG by taking multiple factors into consideration at the same time.

\noindent\textbf{Segment Similarity Graph}.
An untrimmed, long video may contain multiple action instances with large variances due to different scenes, illumination conditions, shooting angles, occlusions, \etc~However, there are always similar motion patterns among multiple instances of the same action category, where some high-quality or easy-to-classify segments recording more complete action instances with less interference provide relatively stable information and low-quality segments can also be complementary to each other. For instance, two temporal segments belonging to the same action class may be occluded in different regions. In this case, one can facilitate the other to perceive the regions that are visible in its own segment. As a result, it is desirable to propagate various kinds of complementary information across all the segments. To this end, we first construct a segment similarity graph by considering the similarities among segment-level features.

Here, we employ the Cosine distance between two original segment-level features to measure their similarity, and construct the similarity graph $\mathcal{G}^{\text{s}}$ by setting the edge weights (\ie, $A^{\text{s}}_{ij}$) between the $i$-th and $j$-th  nodes as follows:
\begin{equation}
A^{\text{s}}_{ij}=\frac{\mathbf{f}_i\cdot \mathbf{f}_j}{||\mathbf{f}_i ||~||\mathbf{f}_j||},
\end{equation}
where $(\cdot)$ is the inner product and $||\cdot||$ is the magnitude.

\noindent\textbf{Temporal Diffusion Graph}.
Since there exist high temporal dependencies across consecutive segments, we also consider the temporal information when constructing the graph. In nature, temporally close segments usually have a high probability of belonging to the same action and tend to have high similarities, \ie, the corresponding edge weights should be relatively large. Moreover, in practice, the temporal convolution in the feature extraction network (\ie, I3D in our experiments) can fuse the temporal information between adjacent segments in a short temporal window. This leads to even higher feature similarities between temporally close segments (\ie, $A^{\text{s}}_{ij}$ tends to be large when $i\rightarrow j$). Therefore, if we construct the temporal graph based on the above facts and add it directly to the segment similarity graph, the propagation of the complementary information is likely to be restricted in a short temporal window and cannot be successfully shared between segments that are far apart. For example, the $i$-th segment $S_i$ containing a high-quality discriminative action instance cannot complement the other inferior instances (belonging to the same action) which are temporally located far away from $S_i$. 

Therefore, we attempt to spread out the complementary information as far as possible so that the discriminability of more segments can be enhanced in the untrimmed, long video, leading to improved localization performance. To this end, we construct a temporal diffusion graph by imposing larger edge weights between farther nodes. Specifically, we construct the temporal diffusion graph $\mathcal{G}^{\text{t}}$ as follows:
\begin{equation}
\label{eq:temporal}
A^{\text{t}}_{ij}=1-\frac{\text{max}(Z-|i-j|,0)}{Z},
\end{equation}
where $Z$ is a hyper-parameter to control the diffusion degree.

\noindent\textbf{Overall Sparse Graph}. By simply combining the two sub-graphs $\mathcal{G}^{\text{s}}$ and $\mathcal{G}^{\text{t}}$, we can obtain our final action complement graph $\mathcal{G}$, of which the adjacency matrix is defined as follows:
\begin{equation}
\mathbf{A}=\frac{\mathbf{A}^{\text{s}}+\alpha \mathbf{A}^{\text{t}}}{2},
\end{equation}
where the two matrices $\mathbf{A}^{\text{s}}$ and $\mathbf{A}^{\text{t}}$ include $A^{\text{s}}_{ij}$ and $A^{\text{t}}_{ij}$ as their ($i,j$)-th entries, respectively, and $\alpha$ is the hyper-parameter for a better trade-off between the two sub-graphs.

Due to that the edge weights of the two sub-graphs are mostly above zero, simply combining them to form the ACG results in a very dense graph. If we directly learn the enhanced features based on this dense graph, we may obtain similar global video-level features for each node/segment since each node is expected to perceive the features of all the rest nodes. This implicitly hinders the discriminability of segment-level features, leading to less accurate localization results. Therefore, it is necessary to make the graph sparse enough to only preserve those most informative nodes. In particular, we set our sparsification criterion based on both a threshold $\lambda$ and a top-$K$ ranking list. The final sparse ACG is constructed as:
\begin{equation}
\label{eq:sparse}
A'_{ij}=\begin{cases}\text{sgn}(A_{ij}-\lambda)\cdot A_{ij}, & \text{rank}_i(j)\le K \\ 0, & \text{rank}_i(j)>K\end{cases}
\end{equation}
where $\text{sgn}(\cdot)$ is an indicator, \ie, $\text{sgn}(x)=1$ if $x>0$; otherwise $\text{sgn}(x)=0$. $\text{rank}_i(j)$ is the ranking number of the $j$-th node w.r.t. the edge weights among all the adjacent nodes of the $i$-th node in the dense graph w.r.t. $\mathbf{A}$. Note that we adopt these two criteria regarding $\lambda$ and $K$ to make the graph sparse, because simply adopting the threshold cannot discard those ambiguous segments in similar scenes but belonging to different action classes. This intuition is also supported by the ablation study in our experiments.

\subsection{Graph Inference}
\label{acgn}

\noindent\textbf{Graph Average}.
After constructing the final sparse ACG, a straightforward way to aggregating all the node-level features is to compute the average features by considering the edge weights as follows:
\begin{equation}
\mathbf{f}^{\text{AVG}}_{i}=\sum_{j=1}^T \hat{A}_{ij} \mathbf{f}_j,
\end{equation}
where $\hat{A}_{ij}$ is the ($i,j$)-th entry of the matrix $\hat{\mathbf{A}}$, which is the row-wise normalized adjacency matrix w.r.t. $\mathbf{A}'$. In practice, we find the averaged feature $\mathbf{f}^{\text{AVG}}_i$ can exchange complementary information to some extent, achieving satisfactory performance as shown in the subsequent experiments.

\noindent\textbf{Graph Convolution}.
In addition to the above average features, we incorporate graph convolutions into our ACGNet to better aggregate node-level features. For a graph convolutional network (GCN) with $M$ layers, the graph convolution operation w.r.t. the $m$-th $(1 \le m \le M)$ layer is as follows:
\begin{equation}
\mathbf{F}^{(m)}=\sigma(\hat{\mathbf{A}} \mathbf{F}^{(m-1)}\mathbf{W}^{(m)}),
\end{equation}
where $\mathbf{F}^{(m)}$ is the feature generated by the $m$-th graph convolutional layer, $\mathbf{F}^{(0)}=\mathbf{F}$ is the original feature, $\mathbf{F}^{\text{GCN}}=\mathbf{F}^{(M)}$ is the final output of the last graph convolutional layer, $\mathbf{W}^{(m)}\in \mathbb{R}^{D\times D}$ is the trainable parameters of the $m$-th layer, and $\sigma(\cdot)$ is the ReLU~\cite{relu} activation function.

Finally, the original features are combined with the graph averaged features and the output features of the GCN to obtain the enhanced discriminative features:
\begin{equation}
\mathbf{F}'=\mathbf{F}+\mathbf{F}^{\text{AVG}}+\mathbf{F}^{\text{GCN}}.
\end{equation}
Since $\mathbf{F}'$ is the enhanced counterpart of the original feature, different WTAL methods can replace their original input by $\mathbf{F}'$, further performing the subsequent localization task.

\subsection{Training Objective}
\label{loss}
To discover the easy-to-classify segments to enhance the features of other similar ones, making more segments easier to be classified, we propose a novel loss based on an `easy positive mining' (EPM) strategy for sufficiently training the joint WTAL network with our ACGNet embedded:
\begin{equation}
\label{eq:loss}
\mathcal{L}_{\text{EPM}}=\frac{1}{N} \sum_{n=1}^N \sum_{i,j=1}^T (p_{n,j} ||\mathbf{f}'_{n,i}-\mathbf{f}_{n,j} ||^2),~\text{s.t.}~A'_{n,ij}>0,
\end{equation}
where $\mathbf{f}'_{n,i}$ is the output feature of the ACGNet w.r.t. the $i$-th segment in the $n$-th video, and $\mathbf{f}_{n,j}$ and $p_{n,j}$ are the original feature and the maximum activation score among all the classes in terms of the $j$-th segment in the same video, respectively.

Based on Eq. (\ref{eq:loss}), the output features of the ACGNet are encouraged to be consistent with the original features of similar segments, especially those `easy positive' examples that can be successfully classified with the highest confidence scores. In other words, the `easy positive' segments can be regarded as the class centroids in the feature space, and we aim to push other similar segments closer to them. Consequently, more action segments become easier to distinguish, finally achieving more accurate location results.

\begin{table*}[t]
	\begin{center}
	\caption{Comparison results on THUMOS'14. * indicates the results based on our implementations.}
	\vspace{-0.2cm}
	\label{THUMOS14}
	\resizebox{0.99\linewidth}{!}{
	\begin{tabular}{l|cccccccc}
	%\begin{tabular}{l|c|C{2cm}|C{2cm}}
	\hline
	\rowcolor{mygray} ~& \multicolumn{8}{|c}{mAP (\%) @ IoU}\\ 
	\rowcolor{mygray}{ } \multirow{-2}{*}{Methods} & \multicolumn{1}{|c}{0.1} & 0.2 & 0.3 & 0.4 & 0.5 & 0.6 & 0.7 & Average \\
	\hline\hline
	TAL-Net~\cite{TAL_Net}&  59.8 & 57.1 & 53.2 & 48.5 & 42.8 & 33.8 & 20.8 & 45.1\\
	BSN~\cite{BSN}&  - & -& 53.5 & 45.0 & 36.9 & 28.4 & 20.0 & - \\
	GTAN~\cite{GTAN}&  69.1 & 63.7 & 57.8 & 47.2 & 38.8 & - & - & - \\
	BMN~\cite{BMN}&  - & -& 56.0 & 47.4 & 38.8 & 29.7 & 20.5 & - \\
	P-GCN~\cite{PGCN}&  69.5 & 67.8& 63.6 & 57.8 & 49.1 & - & - & - \\
	\hline\hline 
    STPN~\cite{STPN}&  52.0 & 44.7    & 35.5  & 25.8  & 16.9  & 9.9   &   4.3 & 27.0\\
    W-TALC~\cite{WTALC}&  55.2 & 49.6    & 40.1  & 31.1  & 22.8  & -     & 7.6  & -\\
    MAAN~\cite{MAAN}&  59.8 & 50.8   & 41.1  & 30.6  & 20.3  & 12.0  & 6.9  & 31.6\\
    Liu \etal~\cite{liu}&  57.4 & 50.8  & 41.2  & 32.1  & 23.1  & 15.0  & 7.0  & 32.4\\
    TSM~\cite{TSM}&  - & -   & 39.5  & -     & 24.5  & -     & 7.1  & -\\
    Nguyen \etal~\cite{Nguyen_etal}&  60.4 & 56.0   & 46.6  & 37.5  & 26.8  & 17.6  & 9.0 & 36.3\\
    RPN~\cite{RPN}&  62.3 & 57.0 & 48.2  & 37.2  & 27.9  & 16.7  & 8.1  & 36.8\\
	Gong \etal~\cite{Gong_etal}&  - & - & 46.9 & 38.9 & 30.1 & 19.8 & 10.4 & - \\
	ActionBytes~\cite{ABytes}&  - & - & 43.0 & 35.8 & 29.0 & - & 9.5 &- \\
	EM-MIL~\cite{EM_MIL}&  59.1 & 52.7 & 45.5 & 36.8 & 30.5 & 22.7 & 16.4 & 37.7 \\
	A2CL-PT~\cite{A2CL}&  61.2 & 56.1 & 48.1 & 39.0 & 30.1 & 19.2 & 10.6 & 37.8 \\
	TSCN~\cite{TSCN}&  63.4 & 57.6 & 47.8 & 37.7 & 28.7 & 19.4 & 10.2 & 37.8 \\
	\hline
    BaS-Net~\cite{BaSNet}&  58.2 & 52.3  & 44.6  & 36.0  & 27.0  & 18.6  & 10.4  & 35.3\\
    BaS-Net*~\cite{BaSNet}&  57.5 & 51.6  & 44.3  & 35.5  & 26.8  & 18.6 & 10.2 & 34.9\\
    \quad+\textbf{ACGNet}	&  \bf{58.8\blue{$^{+1.3}$}} & \bf{53.3\blue{$^{+1.7}$}} & \bf{46.4\blue{$^{+2.1}$}} & \bf{38.3\blue{$^{+2.8}$}} & \bf{29.8\blue{$^{+3.0}$}} & \bf{20.9\blue{$^{+2.3}$}} & \bf{11.2\blue{$^{+1.0}$}} & \bf{37.0\blue{$^{+2.1}$}}\\
	\hline
    DGAM~\cite{DGAM}&  60.0 & 54.2   & 46.8  & 38.2  & 28.8  & 19.8  & 11.4  & 37.0\\
    DGAM*~\cite{DGAM} & 59.4 & 53.4 & 46.1 & 37.0 & 27.8 & 19.5 & 11.0 & 36.3\\
    \quad+\textbf{ACGNet}&  \bf{62.5\blue{$^{+3.1}$}} & \bf{55.9\blue{$^{+2.5}$}} & \bf{48.2\blue{$^{+2.1}$}} & \bf{39.5\blue{$^{+2.5}$}} & \bf{29.6\blue{$^{+1.8}$}} & \bf{20.4\blue{$^{+0.9}$}} & \bf{11.2\blue{$^{+0.2}$}} & \bf{38.2\blue{$^{+1.9}$}}\\
	\hline
    BaM~\cite{BaM}&  67.5 & 61.2 & 52.3 & 43.4 & 33.7 & 22.9 & 12.1 & 41.9\\
    BaM*~\cite{BaM}	& 66.6 & 59.8 & 51.3 & 43.0 & 33.4 & 22.4 & 12.1 & 41.2\\
    \quad+\textbf{ACGNet}&	  \bf{68.1}\blue{$^{+1.5}$} & \bf{62.6}\blue{$^{+2.8}$} & \bf{53.1}\blue{$^{+1.8}$} & \bf{44.6}\blue{$^{+1.6}$} & \bf{34.7}\blue{$^{+1.3}$} & \bf{22.6\blue{$^{+0.2}$}} & \bf{12.0\color{mygray2}{$^{-0.1}$}} & \bf{42.5}\blue{$^{+1.3}$}\\
	\hline
	\end{tabular}
	}
	\end{center}
\vspace{-0.5cm}
\end{table*}

\section{Experiments}
\subsection{Experimental Setup}
\noindent\textbf{Datasets.} \textbf{THUMOS'14}~\cite{THUMOS} contains over 20 hours of videos from 20 sports classes. The dataset is quite challenging due to the diversified video lengths and the large number of action instances ($\sim$15) per video. The training set includes only trimmed videos that are not suitable for action localization, but the validation and test sets provide 200 and 213 untrimmed videos, respectively. Following \cite{BaSNet, DGAM, BaM}, we conduct training on the validation set and perform evaluation on the test set. \textbf{ActivityNet1.2}~\cite{ANet} consists of 100 categories of actions. The training, validation, and test sets are composed of 4,819, 2,383, and 2,480 videos, respectively. However, there are no publicly-available action labels in the test set as it is only used for competitions. Therefore, we follow the general practice in \cite{BaSNet, DGAM, BaM} by employing the training set for training and the validation set for testing.

\textbf{Baselines}. The proposed ACGNet works as a universal module that can be incorporated into different WTAL frameworks. The integration into other frameworks is rather straightforward, and we only need to replace the original features by the enhanced ones obtained by the ACGNet. In our experiments, we adopt three recently proposed WTAL methods, including BaS-Net~\cite{BaSNet}, DGAM~\cite{DGAM}, and BaM~\cite{BaM}.

Due to its flexibility, our ACGNet may also be incorporated into other action detection (\eg, FTAL or the more general spatial-temporal action detection) frameworks. However, it should be noted that ACGNet is to enhance segment-level representations, which is especially crucial in the context of WTAL as only video-level annotations are available. In FTAL or other related tasks, segment-level features can be optimized more easily based on the provided fine-grained (frame-level) annotations.

\noindent\textbf{Evaluation Metrics}.
We adopt the standard metrics for performance evaluation of different methods, \ie, mean Average Precisions (mAPs) under different Intersection of Union (IoU) thresholds. In practice, we adopt the official evaluation code provided by ActivityNet.

\noindent\textbf{Implementation Details}.
The proposed framework is implemented using the PyTorch library. Our ACGNet and the subsequent action localization network are jointly trained in an end-to-end manner. The action localization networks retain the parameter settings in their original papers, and we apply the stochastic gradient descent (SGD) to simultaneously optimize the joint network on an NVIDIA Tesla V100 GPU. For fair comparison with other WTAL methods, we exploit I3D \cite{I3D} to extract the initial segment-level features. The hyper-parameters adopted to construct the ACG are empirically set as follows: $Z=10$, $\alpha=1$, and $\lambda=0.85$. When taking BaS-Net as the action localization network, we set $K$ to 50 and $T=750$ is a fixed value. We set $T=400$ (consistent with the original paper) and $K=T/10=40$ when employing the other two localization frameworks. We utilize a 2-layer graph convolutional network in all the experiments.

\begin{table}[t]
	\begin{center}
	\caption{Comparison results on ActivityNet1.2. * indicates the results based on our implementations.}
	\vspace{-0.2cm}
	\label{ANet}
	\resizebox{0.99\linewidth}{!}{
	\begin{tabular}{l|cccc}
	\hline
	\rowcolor{mygray} ~& \multicolumn{4}{c}{mAP (\%) @ IoU}\\
	\rowcolor{mygray}\multirow{-2}{*}{Methods}& 0.5 & 0.75 & 0.95 & Average \\
	\hline\hline
	Wang \etal~\cite{UntrimmedNets}&	7.4 &	3.2 &	0.7 & 3.6 \\
	AutoLoc~\cite{Autoloc}&	27.3 &	15.1 &	3.3 & 16.0 \\
	W-TALC~\cite{WTALC}	&	37.0 &	12.7 &	1.5 & 18.0 \\
	CleanNet~\cite{CleanNet}&	37.1 &	20.3 &	5.0 & 21.6 \\
	Liu \etal~\cite{liu} &	36.8 &	22.9 &	5.6 & 22.4 \\
	TSM~\cite{TSM}	&	28.3 &	17.0 &	3.5 & 17.1 \\
	RPN~\cite{RPN} &	37.6 &	23.9 &	5.4 & 23.3 \\
Gong \etal~\cite{Gong_etal} & 40.0 & 25.0 & 4.6 & 24.6 \\
EM-MIL~\cite{EM_MIL}  &37.4 & - &- & 20.3 \\
TSCN~\cite{TSCN} &37.6 & 23.7 & 5.7 & 23.6 \\
	\hline
	BaS-Net~\cite{BaSNet}&	38.5 &	24.2 &	5.6 & 24.3 \\
	BaS-Net*~\cite{BaSNet}	&	36.9 &	23.3 &	5.1 & 22.4 \\
	\quad+\textbf{ACGNet}	&	\bf{40.8\blue{$^{+3.9}$}} & \bf{25.3\blue{$^{+2.0}$}} & \bf{5.6\blue{$^{+0.5}$}} & \bf{25.1\blue{$^{+2.7}$}} \\
	\hline
	DGAM~\cite{DGAM}&	41.0 &	23.5 &	5.3 & 24.4 \\
	DGAM*~\cite{DGAM}	&	40.3 &	23.2 &	5.0 & 24.0 \\
	\quad+\textbf{ACGNet}	&	\bf{41.4\blue{$^{+1.1}$}} & \bf{24.2\blue{$^{+1.0}$}} & \bf{5.5\blue{$^{+0.5}$}} & \bf{24.9\blue{$^{+0.9}$}} \\
	\hline
	BaM~\cite{BaM}&	41.2 &	25.6 &	6.0 & 25.9 \\
	BaM*~\cite{BaM}&	40.8 &	24.9 &	5.8 & 25.6 \\
	\quad+\textbf{ACGNet}&	\bf{41.8}\blue{$^{+1.0}$} & \bf{26.0}\blue{$^{+1.1}$} & \bf{5.9\blue{$^{+0.1}$}} & \bf{26.1}\blue{$^{+0.5}$} \\
	\hline
	\end{tabular}
	}
	\end{center}
\vspace{-0.5cm}
\end{table}

\subsection{Comparison to State-of-the-Art Methods}

\textbf{Results on THUMOS'14}. Table~\ref{THUMOS14} shows the localization performance of different methods on THUMOS'14. For fair comparison, we also report the results of the three adopted WTAL frameworks based on our implementations. From the table, we can see that after integrating the proposed ACGNet, the results of the three localization networks are significantly and consistently improved in terms of most IoU thresholds. Notably, when the IoU threshold is set to 0.5, BaS-Net, DGAM, and BaM respectively gain absolute improvements of 3.0\%, 1.8\%, and 1.3\% in mAP. The gain on BaM is not so remarkable, probably due to that BaM greatly improves the discriminability of segment features through background modeling. Such facts indicate the effectiveness of exploiting complementary clues between temporal segments in the weakly-supervised setting. In all, we push the state of the art in WTAL to a large extent, which is even on par with the performance of some fully-supervised approaches.

\noindent\textbf{Results on ActivityNet1.2}. Table~\ref{ANet} shows the comparison results on ActivityNet1.2. Similar to the observations on THUMOS'14, our ACGNet greatly strengthens the existing WTAL frameworks with regard to all the IoU thresholds, and the improvement on BaS-Net is particularly encouraging. Specifically, when adopting 0.5 as the IoU threshold, the mAPs of BaS-Net, DGAM, and BaM are improved by 3.9\%, 1.1\%, and 1.0\%, respectively. This again demonstrates the superiority of the proposed feature enhancement network.

\subsection{Ablation Study}
\label{ablation}
We perform ablation study on BaS-Net as it is the most flexible and efficient among the three baselines. It is worth noting that the number of parameters increases from 26.3 M to 34.6 M when plugging ACGNet into BaS-Net. This complexity is expected as ACGNet includes several processing steps and is not fully optimized. However, considering the flexibility of such a universal module and the consistent performance gains, the increase in complexity is acceptable.

\noindent\textbf{Effects of Graph Design}. We first study the (dis)advantages of different graph designs. Table~\ref{Ab_ACG} shows the results with various degrees of sparsity ($K$). $\mathcal{G}^{1}$ indicates directly using the segment similarity graph $\mathcal{G}^{\text{s}}$; $\mathcal{G}^{2}$ is a variant of our ACG by subtracting the temporal diffusion graph $\mathcal{G}^{\text{t}}$ from $\mathcal{G}^{\text{s}}$; $\mathcal{G}^{3}$ is the proposed ACG. Specifically, when $K$=50, $\mathcal{G}^{3}$ achieves the highest mAP among the three. However, when the graphs become denser, the results tend to decrease gradually. This aligns well with our previous assumption that dense graphs cannot exploit the most discriminative features across all the segments. Finally, we test the performance by using only $\mathcal{G}^{t}$. In this case, most edges in $\mathcal{G}^{t}$ are weighted by one and no meaningful feature enhancement can be guaranteed. Consequently, we only obtain an inferior mAP of 22.1\% when $K$=50.

\begin{table}[t]
	\begin{center}
	\caption{Results of different graph designs on THUMOS'14. * indicates the dense graph without sparsification.}
	\vspace{-0.2cm}
	\label{Ab_ACG}
	\resizebox{0.85\linewidth}{!}{
	\begin{tabular}{c|c|c|c}
	%\begin{tabular}{l|c|C{2cm}|C{2cm}}
	\hline
	\rowcolor{mygray} ~ & \multicolumn{3}{c}{mAP (\%) @ IoU=0.5}\\
	\cline{2-4}

	\rowcolor{mygray}\multirow{-2}{*}{$K$} & { }$\mathcal{G}^{1}=\mathcal{G}^{\text{s}}${ } & $\mathcal{G}^{2}=\mathcal{G}^{\text{s}}-\mathcal{G}^{\text{t}}$ & $\mathcal{G}^{3}=\mathcal{G}^{\text{s}}+\mathcal{G}^{\text{t}}$  \\
	\hline \hline
	1			& 19.3	& 19.7	& 22.2 \\
	5			& 21.8	& 21.6	& 24.3 \\
	20			& 26.8	& 25.9	& 28.1 \\
	50			& \textbf{28.0}	& \textbf{27.2}	& \textbf{29.8} \\
	200			& 27.5	& 26.6	& 28.6 \\
	750*			& 25.9	& 25.1	& 25.3 \\
	\hline
	\end{tabular}
	}
	\end{center}
\vspace{-0.5cm}
\end{table}

\begin{figure*}[!t]
\begin{center}
\includegraphics[width=0.95\linewidth]{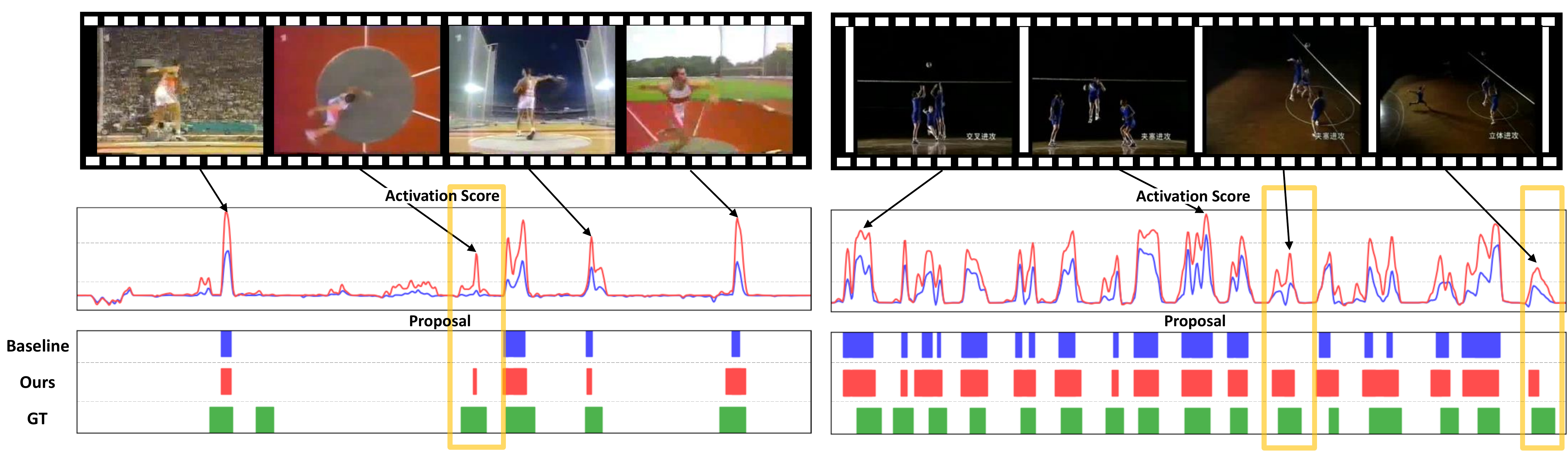}
\vspace{-0.4cm}
\end{center}
   \caption{Qualitative visualization of two typical video examples on THUMOS'14. The results of BaS-Net (Baseline), BaS-Net+ACGNet (Ours), and ground truth (GT) are shown in \blue{blue}, \red{red}, and \green{green}, respectively. The \yellow{yellow} boxes include some difficult cases that Bas-Net fails to detect but can be successfully localized by our method.}
\label{example}
\vspace{-0.3cm}
\end{figure*}

\begin{figure}[!t]
\begin{center}
\includegraphics[width=0.9\linewidth]{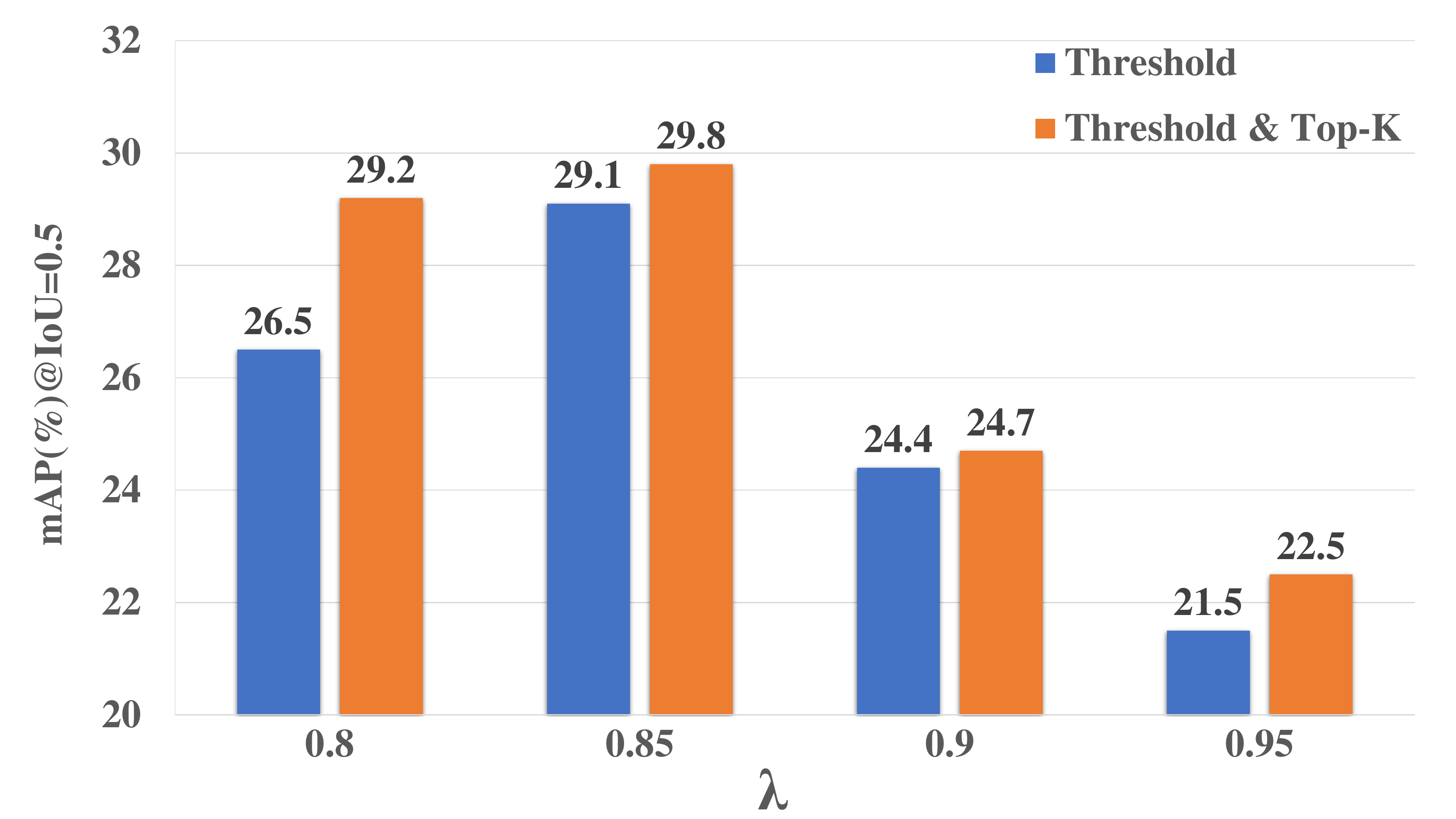}
\vspace{-0.5cm}
\end{center}
   \caption{Comparison results of ACGNet (with BaS-Net) by using different levels of sparsity w.r.t. $\lambda$ on THUMOS'14.}
\label{Ab_Sparse}
\vspace{-0.3cm}
\end{figure}

\noindent\textbf{Effects of Sparsity.} As discussed above, Table \ref{Ab_ACG} includes some results when adopting different sparsity levels w.r.t. $K$. Here, we additionally evaluate how the threshold $\lambda$ affects the final performance. As shown in Figure \ref{Ab_Sparse}, the best results are always achieved by considering both factors (\ie, $\lambda$ and $K$) for graph sparsification. This indicates that simply using a threshold is not enough to maintain the most discriminative nodes. This may be because the scenes remain unchanged in some videos, \ie, the similarity between different segments is always high even if the segments contain different kinds of action instances. In such cases, simply adopting a threshold preserves those irrelevant nodes belonging to distinct categories. By further imposing the top-$K$ constraint, we can remove the ambiguous nodes and keep the most relevant ones, obtaining more discriminative segment-level features.

\noindent\textbf{Component Validation}. Table~\ref{Ab_all} shows the results based on different components in our ACGNet. Concretely, we test the performance when adopting different features and distinct ways of feature combinations. We can see that combining the original features with either the weighted average or the graph convolutional ones can significantly improve the overall accuracy. By fusing all the features, we can achieve the best performance. It is also noteworthy that inferior performance is observed if the EPM loss is discarded during graph training. As mentioned previously, this is because the graph convolutional layers cannot be trained sufficiently. Interestingly, when only using the enhanced graph-based features, the accuracy drops a lot, indicating that taking them as the supplements to the original features shows an effective way to make the potential of the graph features unleashed. In addition, only using $\mathbf{F}^{\text{AVG}}$ performs the worst, as the average features along the whole video cannot represent the distinct temporal dynamics of different action instances.

\begin{table}[!t]
	\begin{center}
	\caption{Results of different components of ACGNet (with BaS-Net) on THUMOS'14.}
	\vspace{-0.2cm}
	\label{Ab_all}
	\resizebox{0.88\linewidth}{!}{
	\begin{tabular}{ccc|c|cc|c}
	\hline
\rowcolor{mygray}\multicolumn{3}{c|}{Feature} & ~ & \multicolumn{2}{c|}{Fusion} & \multicolumn{1}{|c}{mAP (\%) } \\
\rowcolor{mygray}I3D & AVG & GCN & \multirow{-2}{*}{$\mathcal{L}_{\text{EPM}}$} & sum & concat & \multicolumn{1}{|c}{IoU=0.5} \\
\hline\hline
 $\surd$ & & & & & & 26.8\\
 & $\surd$ & & & & & 19.7\\
 $\surd$ & $\surd$ & & & $\surd$ & & 28.5\\
 $\surd$ & $\surd$ & & & & $\surd$ & 29.1\\
 $\surd$ & &  $\surd$& & $\surd$ & & 26.2\\
 $\surd$ & &  $\surd$& & & $\surd$ & 27.3\\
 & & $\surd$ & $\surd$ & & & 22.4\\
 & $\surd$ & $\surd$ & $\surd$ & $\surd$ & & 22.2\\
 $\surd$ & & $\surd$ & $\surd$ & $\surd$ & & 29.0\\
 $\surd$ & & $\surd$ & $\surd$ & & $\surd$ & 29.1\\
 $\surd$ & $\surd$ & $\surd$ & & $\surd$ & & 28.2\\
 $\surd$ & $\surd$ & $\surd$ & $\surd$ & $\surd$ & & \textbf{29.8}\\
 $\surd$ & $\surd$ & $\surd$ & $\surd$ & & $\surd$ & 28.7\\

	\hline
	\end{tabular}
	}
	\end{center}
\vspace{-0.3cm}
\end{table}

\subsection{Qualitative Analysis}
Figure~\ref{example} visualizes some qualitative results. The curves represent the detection activation scores, while the blocks denote the localization results with the IoU threshold at 0.5. It can be observed that most of our scores are higher than the ones delivered by Bas-Net, indicating that our enhanced features are more discriminative for classification. Meanwhile, the scores of other non-action segments remain relatively low, revealing that our method can successfully distinguish action-related segments from irrelevant background. We also note that our detected proposals are more complete, while Bas-Net tends to split one proposal into several individual shorter proposals, leading to degraded accuracy. The difficult cases in the yellow boxes further demonstrate the superiority of our ACGNet.

\section{Conclusion}
This paper presents the ACGNet aiming to enhance the discriminability of segment-level representations of videos for WTAL. The complementary clues from other segments in the same video, particularly the easy-to-classify ones, provides certain supervision to learn more discriminative features. Our ACGNet works as a general module that is flexibly embedded into various existing WTAL frameworks, remarkably boosting the state of the art performance on two challenging benchmarks.

\section{Acknowledgment}
This work is partly supported by the National Natural Science Foundation of China (No. 62022011), the Research Program of State Key Laboratory of Software Development Environment (SKLSDE-2021ZX-04), and the Fundamental Research Funds for the Central Universities.

\small{
\bibliographystyle{ieee_fullname}
\bibliography{aaai22}

\begin{thebibliography}{50}
\providecommand{\natexlab}[1]{#1}

\bibitem[{Abu-El-Haija et~al.(2016)Abu-El-Haija, Kothari, Lee, Natsev,
  Toderici, Varadarajan, and Vijayanarasimhan}]{youtube}
Abu-El-Haija, S.; Kothari, N.; Lee, J.; Natsev, P.; Toderici, G.; Varadarajan,
  B.; and Vijayanarasimhan, S. 2016.
\newblock Youtube-8m: A large-scale video classification benchmark.
\newblock \emph{arXiv preprint arXiv:1609.08675}.

\bibitem[{Caba~Heilbron et~al.(2015)Caba~Heilbron, Escorcia, Ghanem, and
  Carlos~Niebles}]{ANet}
Caba~Heilbron, F.; Escorcia, V.; Ghanem, B.; and Carlos~Niebles, J. 2015.
\newblock Activitynet: A large-scale video benchmark for human activity
  understanding.
\newblock In \emph{CVPR}.

\bibitem[{Carreira and Zisserman(2017)}]{I3D}
Carreira, J.; and Zisserman, A. 2017.
\newblock Quo Vadis, Action Recognition? A New Model and the Kinetics Dataset.
\newblock In \emph{CVPR}.

\bibitem[{Chao et~al.(2018)Chao, Vijayanarasimhan, Seybold, Ross, Deng, and
  Sukthankar}]{TAL_Net}
Chao, Y.-W.; Vijayanarasimhan, S.; Seybold, B.; Ross, D.~A.; Deng, J.; and
  Sukthankar, R. 2018.
\newblock Rethinking the faster r-cnn architecture for temporal action
  localization.
\newblock In \emph{CVPR}.

\bibitem[{Escorcia et~al.(2016)Escorcia, Heilbron, Niebles, and
  Ghanem}]{Victor2016DAPs}
Escorcia, V.; Heilbron, F.~C.; Niebles, J.~C.; and Ghanem, B. 2016.
\newblock DAPs: Deep Action Proposals for Action Understanding.
\newblock In \emph{ECCV}.

\bibitem[{Feichtenhofer et~al.(2019)Feichtenhofer, Fan, Malik, and
  He}]{SlowFast}
Feichtenhofer, C.; Fan, H.; Malik, J.; and He, K. 2019.
\newblock SlowFast Networks for Video Recognition.
\newblock In \emph{ICCV}.

\bibitem[{Gong et~al.(2020)Gong, Wang, Mu, and Tian}]{Gong_etal}
Gong, G.; Wang, X.; Mu, Y.; and Tian, Q. 2020.
\newblock Learning Temporal Co-Attention Models for Unsupervised Video Action
  Localization.
\newblock In \emph{CVPR}.

\bibitem[{Huang et~al.(2020)Huang, Huang, Ouyang, Wang et~al.}]{RPN}
Huang, L.; Huang, Y.; Ouyang, W.; Wang, L.; et~al. 2020.
\newblock Relational Prototypical Network for Weakly Supervised Temporal Action
  Localization.
\newblock In \emph{AAAI}.

\bibitem[{Huang, Sugano, and Sato(2020)}]{GTRM}
Huang, Y.; Sugano, Y.; and Sato, Y. 2020.
\newblock Improving Action Segmentation via Graph-Based Temporal Reasoning.
\newblock In \emph{CVPR}.

\bibitem[{Idrees et~al.(2017)Idrees, Zamir, Jiang, Gorban, Laptev, Sukthankar,
  and Shah}]{THUMOS}
Idrees, H.; Zamir, A.~R.; Jiang, Y.-G.; Gorban, A.; Laptev, I.; Sukthankar, R.;
  and Shah, M. 2017.
\newblock The THUMOS challenge on action recognition for videos “in the
  wild”.
\newblock \emph{CVIU}, 155: 1--23.

\bibitem[{Jain, Ghodrati, and Snoek(2020)}]{ABytes}
Jain, M.; Ghodrati, A.; and Snoek, C. G.~M. 2020.
\newblock ActionBytes: Learning From Trimmed Videos to Localize Actions.
\newblock In \emph{CVPR}.

\bibitem[{Kong et~al.(2020)Kong, Huang, Qin, and Wang}]{sports}
Kong, L.; Huang, D.; Qin, J.; and Wang, Y. 2020.
\newblock A Joint Framework for Athlete Tracking and Action Recognition in
  Sports Videos.
\newblock \emph{IEEE TCSVT}, 30(2): 532--548.

\bibitem[{Lee, Uh, and Byun(2020)}]{BaSNet}
Lee, P.; Uh, Y.; and Byun, H. 2020.
\newblock Background Suppression Network for Weakly-Supervised Temporal Action
  Localization.
\newblock In \emph{AAAI}.

\bibitem[{Lee et~al.(2021)Lee, Wang, Lu, and Byun}]{BaM}
Lee, P.; Wang, J.; Lu, Y.; and Byun, H. 2021.
\newblock Background Modeling via Uncertainty Estimation for Weakly-supervised
  Action Localization.
\newblock In \emph{AAAI}.

\bibitem[{Li et~al.(2019)Li, Yao, Qiu, Li, and Mei}]{li2019long}
Li, D.; Yao, T.; Qiu, Z.; Li, H.; and Mei, T. 2019.
\newblock Long Short-Term Relation Networks for Video Action Detection.
\newblock In \emph{ACM MM}.

\bibitem[{Li et~al.(2020)Li, Ji, Shi, Zhang, Kang, and Wang}]{TEA}
Li, Y.; Ji, B.; Shi, X.; Zhang, J.; Kang, B.; and Wang, L. 2020.
\newblock TEA: Temporal Excitation and Aggregation for Action Recognition.
\newblock In \emph{CVPR}.

\bibitem[{Lin et~al.(2019)Lin, Liu, Li, Ding, and Wen}]{BMN}
Lin, T.; Liu, X.; Li, X.; Ding, E.; and Wen, S. 2019.
\newblock BMN: Boundary-matching network for temporal action proposal
  generation.
\newblock In \emph{ICCV}.

\bibitem[{Lin, Zhao, and Shou(2017)}]{LinSingle}
Lin, T.; Zhao, X.; and Shou, Z. 2017.
\newblock Single Shot Temporal Action Detection.
\newblock In \emph{ACM MM}.

\bibitem[{Lin et~al.(2018)Lin, Zhao, Su, Wang, and Yang}]{BSN}
Lin, T.; Zhao, X.; Su, H.; Wang, C.; and Yang, M. 2018.
\newblock BSN: Boundary Sensitive Network for Temporal Action Proposal
  Generation.
\newblock In \emph{ECCV}.

\bibitem[{Liu, Jiang, and Wang(2019)}]{liu}
Liu, D.; Jiang, T.; and Wang, Y. 2019.
\newblock Completeness modeling and context separation for weakly supervised
  temporal action localization.
\newblock In \emph{CVPR}.

\bibitem[{Liu et~al.(2020)Liu, Luo, Wang, Wang, and Lu}]{TEINet}
Liu, Z.; Luo, D.; Wang, Y.; Wang, L.; and Lu, T. 2020.
\newblock TEINet: Towards an Efficient Architecture for Video Recognition.
\newblock In \emph{AAAI}.

\bibitem[{Liu et~al.(2019)Liu, Wang, Zhang, Gao, Niu, Zheng, and
  Hua}]{CleanNet}
Liu, Z.; Wang, L.; Zhang, Q.; Gao, Z.; Niu, Z.; Zheng, N.; and Hua, G. 2019.
\newblock Weakly supervised temporal action localization through contrast based
  evaluation networks.
\newblock In \emph{ICCV}.

\bibitem[{Long et~al.(2019)Long, Yao, Qiu, Tian, Luo, and Mei}]{GTAN}
Long, F.; Yao, T.; Qiu, Z.; Tian, X.; Luo, J.; and Mei, T. 2019.
\newblock Gaussian temporal awareness networks for action localization.
\newblock In \emph{CVPR}.

\bibitem[{Luo et~al.(2020)Luo, Guillory, Shi, Ke, Wan, Darrell, and
  Xu}]{EM_MIL}
Luo, Z.; Guillory, D.; Shi, B.; Ke, W.; Wan, F.; Darrell, T.; and Xu, H. 2020.
\newblock Weakly-Supervised Action Localization with Expectation-Maximization
  Multi-Instance Learning.
\newblock In Vedaldi, A.; Bischof, H.; Brox, T.; and Frahm, J.-M., eds.,
  \emph{ECCV}.

\bibitem[{Min and Corso(2020)}]{A2CL}
Min, K.; and Corso, J.~J. 2020.
\newblock Adversarial Background-Aware Loss for Weakly-Supervised Temporal
  Activity Localization.
\newblock In Vedaldi, A.; Bischof, H.; Brox, T.; and Frahm, J.-M., eds.,
  \emph{ECCV}.

\bibitem[{Nair and Hinton(2010)}]{relu}
Nair, V.; and Hinton, G.~E. 2010.
\newblock Rectified Linear Units Improve Restricted Boltzmann Machines Vinod
  Nair.
\newblock In \emph{ICML}.

\bibitem[{Nguyen et~al.(2018)Nguyen, Liu, Prasad, and Han}]{STPN}
Nguyen, P.; Liu, T.; Prasad, G.; and Han, B. 2018.
\newblock Weakly supervised action localization by sparse temporal pooling
  network.
\newblock In \emph{CVPR}.

\bibitem[{Nguyen, Ramanan, and Fowlkes(2019)}]{Nguyen_etal}
Nguyen, P.~X.; Ramanan, D.; and Fowlkes, C.~C. 2019.
\newblock Weakly-supervised action localization with background modeling.
\newblock In \emph{ICCV}.

\bibitem[{Ni, Qin, and Huang(2021)}]{igmn}
Ni, J.; Qin, J.; and Huang, D. 2021.
\newblock Identity-Aware Graph Memory Network for Action Detection.
\newblock In \emph{ACM MM}.

\bibitem[{Paul, Roy, and Roy-Chowdhury(2018)}]{WTALC}
Paul, S.; Roy, S.; and Roy-Chowdhury, A.~K. 2018.
\newblock W-talc: Weakly-supervised temporal activity localization and
  classification.
\newblock In \emph{ECCV}.

\bibitem[{Qi et~al.(2020)Qi, Wang, Qin, Li, Luo, and Van~Gool}]{stagnet}
Qi, M.; Wang, Y.; Qin, J.; Li, A.; Luo, J.; and Van~Gool, L. 2020.
\newblock stagNet: An Attentive Semantic RNN for Group Activity and Individual
  Action Recognition.
\newblock \emph{IEEE TCSVT}, 30(2): 549--565.

\bibitem[{Qin et~al.(2017)Qin, Liu, Shao, Shen, Ni, Chen, and Wang}]{zsecoc}
Qin, J.; Liu, L.; Shao, L.; Shen, F.; Ni, B.; Chen, J.; and Wang, Y. 2017.
\newblock Zero-Shot Action Recognition with Error-Correcting Output Codes.
\newblock In \emph{CVPR}.

\bibitem[{Rashid, Kjellstrm, and Yong(2020)}]{Action_Graph}
Rashid, M.; Kjellstrm, H.; and Yong, J.~L. 2020.
\newblock Action Graphs: Weakly-supervised Action Localization with Graph
  Convolution Networks.
\newblock In \emph{WACV}.

\bibitem[{Shi et~al.(2020)Shi, Dai, Mu, and Wang}]{DGAM}
Shi, B.; Dai, Q.; Mu, Y.; and Wang, J. 2020.
\newblock Weakly-Supervised Action Localization by Generative Attention
  Modeling.
\newblock In \emph{CVPR}.

\bibitem[{Shou et~al.(2017)Shou, Chan, Zareian, Miyazawa, and Chang}]{CDC}
Shou, Z.; Chan, J.; Zareian, A.; Miyazawa, K.; and Chang, S.-F. 2017.
\newblock CDC: Convolutional-De-Convolutional Networks for Precise Temporal
  Action Localization in Untrimmed Videos.
\newblock In \emph{CVPR}.

\bibitem[{Shou et~al.(2018)Shou, Gao, Zhang, Miyazawa, and Chang}]{Autoloc}
Shou, Z.; Gao, H.; Zhang, L.; Miyazawa, K.; and Chang, S.-F. 2018.
\newblock Autoloc: Weakly-supervised temporal action localization in untrimmed
  videos.
\newblock In \emph{ECCV}.

\bibitem[{Wang et~al.(2017)Wang, Xiong, Lin, and Van~Gool}]{UntrimmedNets}
Wang, L.; Xiong, Y.; Lin, D.; and Van~Gool, L. 2017.
\newblock Untrimmednets for weakly supervised action recognition and detection.
\newblock In \emph{CVPR}.

\bibitem[{Wang et~al.(2020)Wang, Gao, Wang, Li, and Wu}]{BCN}
Wang, Z.; Gao, Z.; Wang, L.; Li, Z.; and Wu, G. 2020.
\newblock Boundary-Aware Cascade Networks for Temporal Action Segmentation.
\newblock In Vedaldi, A.; Bischof, H.; Brox, T.; and Frahm, J.-M., eds.,
  \emph{ECCV}.

\bibitem[{Wu et~al.(2019)Wu, Feichtenhofer, Fan, He, Krahenbuhl, and
  Girshick}]{LFB}
Wu, C.-Y.; Feichtenhofer, C.; Fan, H.; He, K.; Krahenbuhl, P.; and Girshick, R.
  2019.
\newblock Long-term feature banks for detailed video understanding.
\newblock In \emph{CVPR}.

\bibitem[{Wu et~al.(2020)Wu, Kuang, Wang, Zhang, and Wu}]{CRCNN}
Wu, J.; Kuang, Z.; Wang, L.; Zhang, W.; and Wu, G. 2020.
\newblock Context-Aware RCNN: A Baseline for Action Detection in Videos.
\newblock In Vedaldi, A.; Bischof, H.; Brox, T.; and Frahm, J.-M., eds.,
  \emph{ECCV}.

\bibitem[{Xu et~al.(2020)Xu, Zhao, Rojas, Thabet, and Ghanem}]{G_TAD}
Xu, M.; Zhao, C.; Rojas, D.~S.; Thabet, A.; and Ghanem, B. 2020.
\newblock G-TAD: Sub-Graph Localization for Temporal Action Detection.
\newblock In \emph{CVPR}.

\bibitem[{Yang et~al.(2019)Yang, Yang, Liu, Xiao, Davis, and Kautz}]{STEP}
Yang, X.; Yang, X.; Liu, M.-Y.; Xiao, F.; Davis, L.~S.; and Kautz, J. 2019.
\newblock STEP: Spatio-Temporal Progressive Learning for Video Action
  Detection.
\newblock In \emph{CVPR}.

\bibitem[{Yang et~al.(2021)Yang, Huang, Qin, and Wang}]{icassp}
Yang, Z.; Huang, D.; Qin, J.; and Wang, Y. 2021.
\newblock Human-Aware Coarse-to-Fine Online Action Detection.
\newblock In \emph{ICASSP}.

\bibitem[{Yu et~al.(2019)Yu, Ren, Li, Yan, Xu, and Yuan}]{TSM}
Yu, T.; Ren, Z.; Li, Y.; Yan, E.; Xu, N.; and Yuan, J. 2019.
\newblock Temporal structure mining for weakly supervised action detection.
\newblock In \emph{ICCV}.

\bibitem[{Yuan et~al.(2019)Yuan, Lyu, Shen, Tsang, and Yeung}]{MAAN}
Yuan, Y.; Lyu, Y.; Shen, X.; Tsang, I.~W.; and Yeung, D.-Y. 2019.
\newblock Marginalized average attentional network for weakly-supervised
  learning.
\newblock In \emph{ICLR}.

\bibitem[{Zeng et~al.(2019)Zeng, Huang, Tan, Rong, Zhao, Huang, and Gan}]{PGCN}
Zeng, R.; Huang, W.; Tan, M.; Rong, Y.; Zhao, P.; Huang, J.; and Gan, C. 2019.
\newblock Graph convolutional networks for temporal action localization.
\newblock In \emph{ICCV}.

\bibitem[{Zhai et~al.(2020)Zhai, Wang, Tang, Zhang, Yuan, and Hua}]{TSCN}
Zhai, Y.; Wang, L.; Tang, W.; Zhang, Q.; Yuan, J.; and Hua, G. 2020.
\newblock Two-Stream Consensus Network for Weakly-Supervised Temporal
  Action Localization.
\newblock In Vedaldi, A.; Bischof, H.; Brox, T.; and Frahm, J.-M., eds.,
  \emph{ECCV}.

\bibitem[{Zhang et~al.(2019)Zhang, Xu, Cheng, Niu, Pu, Wu, and
  Zou}]{zhang2019adversarial}
Zhang, C.; Xu, Y.; Cheng, Z.; Niu, Y.; Pu, S.; Wu, F.; and Zou, F. 2019.
\newblock Adversarial Seeded Sequence Growing for Weakly-Supervised Temporal
  Action Localization.
\newblock In \emph{ACM MM}.

\bibitem[{Zhao et~al.(2017)Zhao, Xiong, Wang, Wu, Tang, and Lin}]{SSN}
Zhao, Y.; Xiong, Y.; Wang, L.; Wu, Z.; Tang, X.; and Lin, D. 2017.
\newblock Temporal Action Detection With Structured Segment Networks.
\newblock In \emph{ICCV}.

\bibitem[{Zolfaghari, Singh, and Brox(2018)}]{ECO}
Zolfaghari, M.; Singh, K.; and Brox, T. 2018.
\newblock ECO: Efficient convolutional network for online video understanding.
\newblock In \emph{ECCV}.

\end{thebibliography}
}

\end{document}